\documentclass{article}

\PassOptionsToPackage{numbers, compress}{natbib}

      \usepackage[preprint]{neurips_2020}

\usepackage[utf8]{inputenc} 
\usepackage[T1]{fontenc}    
\usepackage{hyperref}       
\usepackage{url}            
\usepackage{booktabs}       
\usepackage{amsfonts}       
\usepackage{nicefrac}       
\usepackage{microtype}      
\usepackage{graphicx}
\graphicspath{ {figures/} }
\usepackage{caption}
\usepackage{subcaption}
\usepackage{float}
\usepackage{amsmath}
\DeclareMathOperator*{\argmin}{arg\,min}
\usepackage{cleveref}
\newsavebox{\imagebox}

\title{Na\"{i}ve Artificial Intelligence}

\author{%
  Tomer Barak\\
  The Edmond and Lily Safra Center for Brain Sciences\\
  The Hebrew University, Jerusalem \\
  \texttt{tomer.barak@mail.huji.ac.il}
  \And
  Yehonatan Avidan\\
  The Edmond and Lily Safra Center for Brain Sciences\\
  The Hebrew University, Jerusalem\\
  \texttt{yehonatan.avidan@gmail.com}
  \And
  Yonatan Loewestein\\
    Department of Cognitive Sciences\\
The Federmann Center for the Study of Rationality\\
The Alexander Silberman Institute of Life Sciences\\
The Edmond and Lily Safra Center for Brain Sciences\\
  The Hebrew University, Jerusalem\\
  \texttt{yonatan.loewenstein@mail.huji.ac.il}
}

\begin{document}

\maketitle

\begin{abstract}
In the cognitive sciences, it is common to distinguish between crystal intelligence, the ability to utilize knowledge acquired through past learning or experience and fluid intelligence, the ability to solve novel problems without relying on prior knowledge. Using this cognitive distinction between the two types of intelligence, extensively-trained deep networks that can play chess or Go exhibit crystal but not fluid intelligence. In humans, fluid intelligence is typically studied and quantified using intelligence tests. Previous studies have shown that deep networks can solve some forms of intelligence tests, but only after extensive training. Here we present a computational model that solves intelligence tests without any prior training. This ability is based on continual inductive reasoning, and is implemented by deep unsupervised latent-prediction networks. Our work demonstrates the potential fluid intelligence of deep networks. Finally, we propose that the computational principles underlying our approach can be used to model fluid intelligence in the cognitive sciences.
\end{abstract}

\section{Introduction}

\begin{figure}[h]
\centering
  \begin{subfigure}[b]{0.7\textwidth}
    \includegraphics[width=\textwidth]{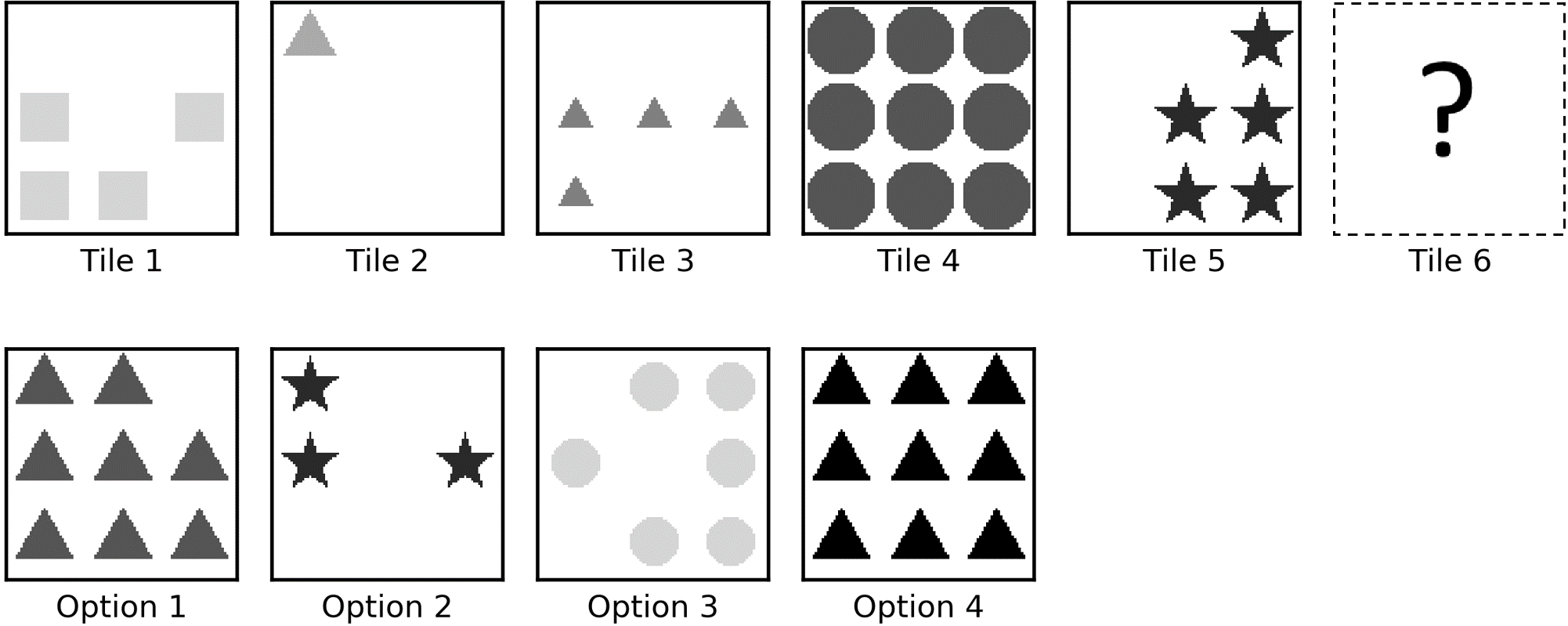}
  \end{subfigure}
  \hfill
  \begin{subfigure}[b]{0.8\textwidth}
    \includegraphics[width=\textwidth]{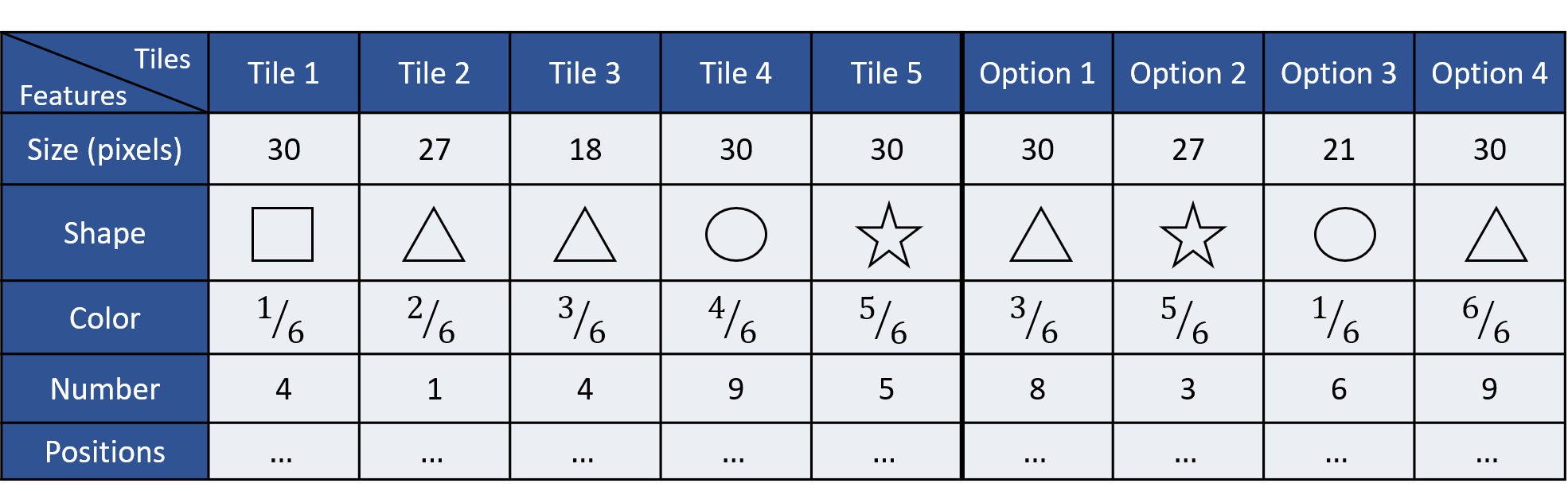}
  \end{subfigure}
  \caption{\textbf{Intelligence tests for measuring inductive reasoning}. In each test, a sequence of $t=5$ tiles is presented, and the objective is to choose the next tile from a set of $n=4$ alternatives. The tiles are 100$\times$100 pixels images that are characterized by the features: \{\textit{Color, Number, Shape, Size, Positions}\}. One of these features follows a rule. In this example, the color intensity increases along the sequence. The other features can be either constant or change randomly. When a feature changes at random we refer to it as a distractor and the difficulty of a tests is defined by the number of distractors. In this example, the number, shape, size and positions of the objects are all distractors. The tests were constructed according to \cite{Wang2015,Barrett2018}, focusing on the settings that are best for measuring inductive reasoning (see supplementary materials for more details).}
  \label{fig:test_example}
\end{figure}

Consider the intelligence test depicted in Fig.\ref{fig:test_example}: five ordered tiles are presented to the agent in a one-dimensional Raven's Progressive Matrix (RPM) test. The tiles are characterized by features: the number of objects, their color, shape, size, and positions. One of the features changes in accordance with a predefined rule. The objective of the agent is to select the sixth tile that adheres to that rule out of a selection of four alternative tiles. To complicate the task, randomly-changing features, which we refer to as \textit{distractors}, also characterize the tiles. The task is challenging because the relevant feature and rule should be simultaneously inferred from the examples. Solving RPMs requires \textit{inductive reasoning}, loosely defined as the ability to derive general rules out of specific observations \cite{Blum1975,Sternberg1983,Siebers2015}. Inductive reasoning is arguably the most important component of fluid intelligence, a cognitive faculty that is correlated with skills such as problem solving and comprehension. Indeed, RPMs are commonly used to quantify fluid intelligence\footnote{In many cases, tiles in RPM tests are arranged in a 3$\times$3 matrix, which requires disassembling the problem to multiple sub-goals of finding the features and rule of each single row and column \cite{Carpenter1990}.} \cite{Kaplan2009}.

Incorporating inductive reasoning in machines has been challenging. Traditional computational models that solved RPMs utilized either a set of predefined features \cite{Rasmussen2011}, a set of predefined rules \cite{Sun2018} or both \cite{Carpenter1990}. With the development of modern machine learning, the use of predefined features or rules became unnecessary, as they can be learned by deep artificial neural networks. However, such learning requires extensive supervised learning \cite{Barrett2018, Hill2019}. By contrast, humans effectively perform inductive reasoning as an unsupervised continual process \cite{Siebers2015} using a small number of examples \cite{Barascud2016}.

Here we present a prediction model that performs inductive reasoning without prior training. The model is based on an unsupervised latent-prediction network \cite{Oord2018,Anand2019,Yan2020}, which means that it looks for a predictable latent representation rather than attempting to make predictions in pixel space. Thus, the model can find a latent representation that corresponds to the predictable feature and its underlying rule. We show that this enables the model to solve RPMs, hence to perform inductive reasoning.



\section{Inductive Reasoning Model}

\label{sec:ind_res_model}

\paragraph{The challenge.} 
In the test depicted in Fig. 1, the world generates a sequence of grayscale images $\mathbf{x}^j$ in the following way: each image is characterized by a low dimensional vector of features, $\mathbf{f}^j$ where $f_i^j$ denotes the value of feature $i$ in image $j$. The image $\mathbf{x}^j$ is constructed by applying a non-linear and complex generative function from the low features dimension to the high pixel space $\mathbf{x}^j=g\left(\mathbf{f}^j\right)$. Importantly, while all features but one are either constant over the images or are i.i.d., one of the features $f^j_p$ changes predictably according to a specific rule. After observing a sequence of $t$ images, the agent's task is to select the correct $t+1^{\text{th}}$ image from a set of $n$ images that were generated using the same generative model over the low features dimension. In the correct image, $f^{t+1}_p$ follows the predictable rule whereas it is randomly chosen for the incorrect images. This task is difficult because neither the features (or even the set of possible features) nor the rule (or even the set of possible rules) are given to the agent, and they have to be inferred from the sequence of $t$ images.  

\paragraph{Predictable representations.} 

In the cognitive sciences literature, it has been shown that humans solve intelligence tests by concurrently identifying the features in the sequence of images and the rules that underlie their change \cite{Sternberg1983}. The relationships between the $\mathbf{x}^t$ image and the $n$ alternative tiles are then considered in view of the identified features and rules. The selected image is the one that is the most congruent with the rule (\cite{Sternberg1983, Carpenter1990}). Motivated by this solution, we sought to construct a network that concurrently identifies the predictable feature and the rule. Rather than attempting to predict the $t+1^{\text{th}}$ image, it predicts its lower-dimensional representation in the feature space. The disadvantage of this approach is that the feature needs to be inferred from the images. However, the advantages of making predictions in the latent feature space are (1) its dimensionality is much smaller than that of the pixel space, which implies that fewer images are needed for making such a prediction. (2) Some of the irrelevant features may be stochastic. Thus, the images in the pixel-space may not be predictable. 

By construction, there exists a function $Z^*$ from the image dimension to a scalar that extracts the relevant feature from an image $f_p^j=Z^*(\mathbf{x}^j)$ and a function $T^*$ that describes the rule, such that $T^*\left(f_p^j\right)=f_p^{j+1}$. $Z^*$ and $T^*$ are solutions to the equation

\begin{equation}\label{eq:prediction}
    T(Z(\mathbf{x}^j))=Z(\mathbf{x}^{j+1})
\end{equation}

From a cognitive point of view, the function $Z$ is an \textit{encoder} that projects the image to a one-dimensional variable, and the function $T$ is a \textit{predictor} that predicts the value of the projection of the image in the next tile based on that projection in the current tile. 

\paragraph{Dynamic representations.} Naively, by solving equation (\ref{eq:prediction}) an agent can extract the solutions $Z^*$ and $T^*$ and use them to make predictions. However, there is no unique solution to equation (\ref{eq:prediction}), and not all solutions to this equation are useful for solving the task. One trivial solution to equation (\ref{eq:prediction}) is: $Z(\mathbf{x}^j)=Z(\mathbf{x}^{j+1})$ $\forall{j}$ and $T\left(Z\right)=Z$. This solution is clearly not useful for selecting the $t+1^{\text{th}}$ image. Therefore, in order to make predictions we should seek a dynamic solution that satisfies the inequality 

\begin{equation} \label{eq:disentangle}
    Z(\mathbf{x}^j)\neq Z(\mathbf{x}^{j+1})
\end{equation}

\paragraph{Bounded representations.} Finally, for every solution $Z$ and $T$ to equations (\ref{eq:prediction}) and (\ref{eq:disentangle}), there is a continuum of other solutions that are given by the stretching and / or the shifting of the function $Z$ with a corresponding compensation of the function $Z$. It is possible to set the scale of the solutions by bounding the representations, e.g., to be between $-1$ and $1$:

\begin{equation} \label{eq:bound}
    \max_{j}{\left|Z(\mathbf{x}^j)\right|}\leq 1
\end{equation}

From here on, $Z^*$ and $T^*$ will denote the set of all stretched and shifted $Z$ and $T$ such that the $Z$-s are in the homeomorphism class of the ground truth $Z^*$ and the $T$-s are the corresponding predictors, $T(Z(\mathbf{x}))=T^*(Z^*(\mathbf{x}))$ $\forall{\mathbf{x}}$.

\paragraph{Decision making.} 

\begin{figure}[h]
  \centering
  \includegraphics[width=0.7\textwidth]{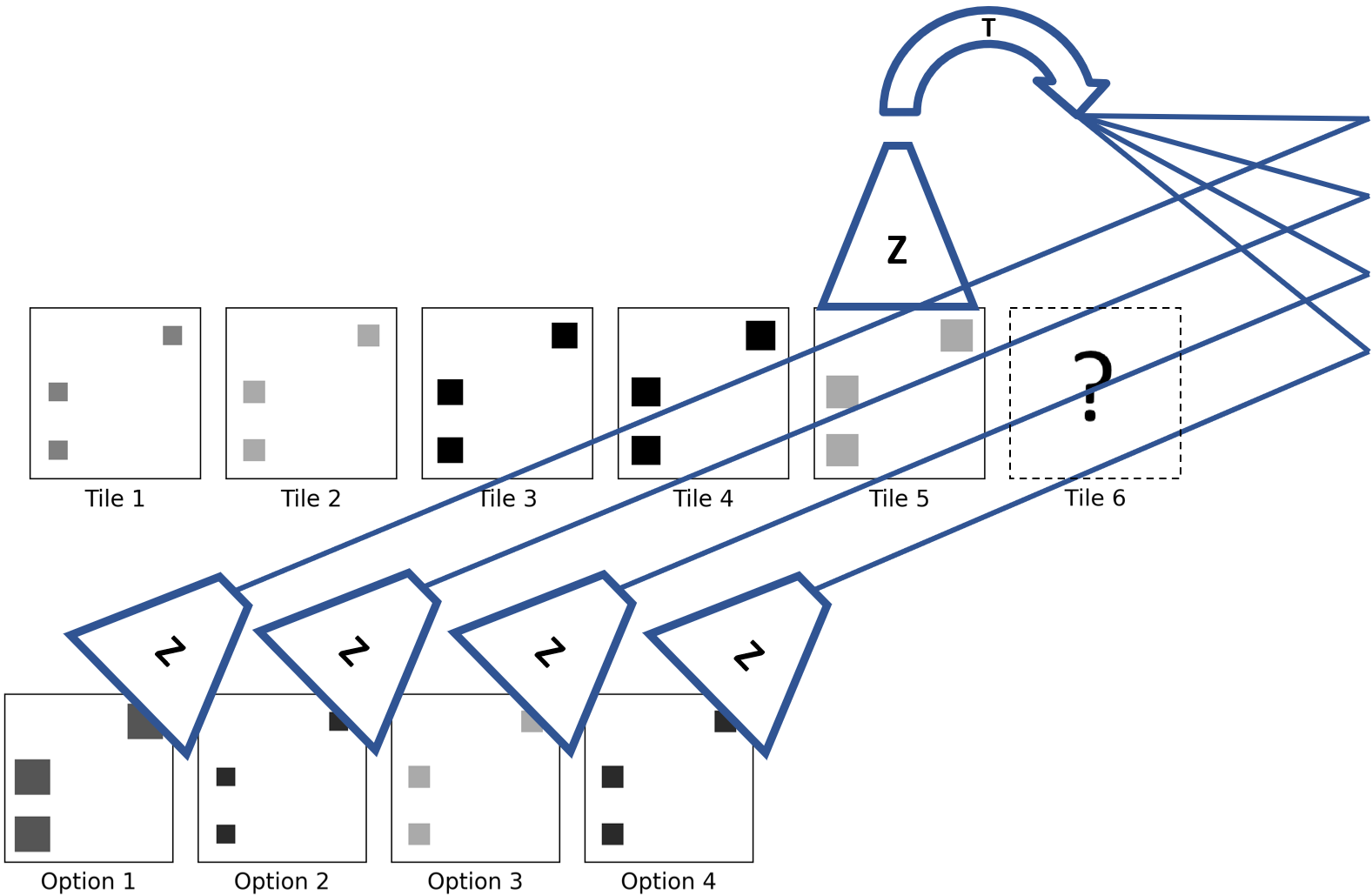}
  \caption{\textbf{Inductive reasoning model for solving intelligence tests}. The model is composed of two functions: An \textit{encoder} $Z$ that encodes the relevant feature, and a \textit{predictor} $T$ that predicts in latent space. Decision is made by encoding the image of the last test tile $\mathbf{x}^t$ and determining which of the options $\mathbf{o}^k$ is best predicted in the latent space. In this example, a good encoder will encode the size of the squares, and the predictor will predict that they monotonically increase - together determining that Option 1 best completes the sequence.}
  \label{fig:decision_making}
\end{figure}

Given $Z=Z^*$ and $T=T^*$, choosing the correct tile is trivial. The correct image $\mathbf{o}^{\text{correct}}$ satisfies $T^*(Z^*(\mathbf{x}^t))=Z^*(\mathbf{o}^{\text{correct}})$ and therefore, 

\begin{equation} \label{eq:decision}
    \text{Correct option} = \argmin_k \left( T(Z(\mathbf{x}^t))-Z(\mathbf{o}^k) \right) ^2
\end{equation}

where the alternatives are denoted by $\{ \mathbf{o}^k\}$ (Fig. \ref{fig:decision_making}).

\paragraph{Inductive reasoning as an optimization problem.} The exact solution, $Z^*$ and $T^*$ satisfies two conditions, Eqs. (\ref{eq:prediction}) and (\ref{eq:disentangle}). Eq. (\ref{eq:bound}) sets the scale of the solution. We propose that good encoder and predictor can be found by minimizing the three loss functions that correspond to the three equations (adapted from \cite{Francois-Lavet2019}):

\begin{equation}\label{eq:losses}
\begin{split}
\mathcal{L}_{pred}=\left( T \left( Z \left( \mathbf{x}^j \right) \right) -Z \left( \mathbf{x}^{j+1} \right) \right) ^2 \\[1ex]
\mathcal{L}_{dis}=\exp{\left(-\frac{\left|Z\left(\mathbf{x}^j\right)-Z\left(\mathbf{x}^{j+1}\right)\right|}{\sigma}\right)} \\[1ex]
\mathcal{L}_{bound}=\max\left ( \max_j{\left ( Z(\mathbf{x}^j)^2-1 \right )},0\right )
\end{split}
\end{equation}
    
The loss function $\mathcal{L}_{pred}$ is minimized when equation (\ref{eq:prediction}) is satisfied, i.e., when a predictable low-dimensional representation is found. The second loss function, $\mathcal{L}_{dis}$, decreases the more dynamic the representations are. The parameter $\sigma$ determines the scale of difference between consecutive representations. Note that because of the exponential shape of the loss function, if the representations are sufficiently different (relative to $\sigma$) then further separating them will have only a small effect on the loss function. In our simulations we used $\sigma=0.2$. This parameter becomes meaningful in view of $\mathcal{L}_{bound}$, which acts to maintain the representations in the $[-1,1]$ range.

\section{Results}
\label{sec:results}

\paragraph{The network model.} 
For the encoder $Z(\mathbf{x})$ we used a 8-layer convolutional neural network from the 100$\times$100 pixel space to a single neuron. The predictor $T(Z)$ calculates the representations transition $T(Z)=Z+\Delta T(Z)$ where $\Delta T$ is a 5-layer fully-connected network. To learn the parameters of the two networks, we concurrently minimized the loss functions, Eq. (\ref{eq:losses}), each with its own RMSprop optimizer \cite{Francois-Lavet2019}. Given a sequence of $t$ tiles $\{\mathbf{x}^j\}_{j=1}^{t}$, each optimization step optimizes a minibatch that consists of the $t-1$ consecutive pairs of tiles (see supplementary materials for more information).

\subsection{The expressivity of the model - extensive training}
\label{sec:extensive_trainig}

\begin{figure}[H]
  \begin{subfigure}[b]{0.45\textwidth} 
    \includegraphics[width=\textwidth]{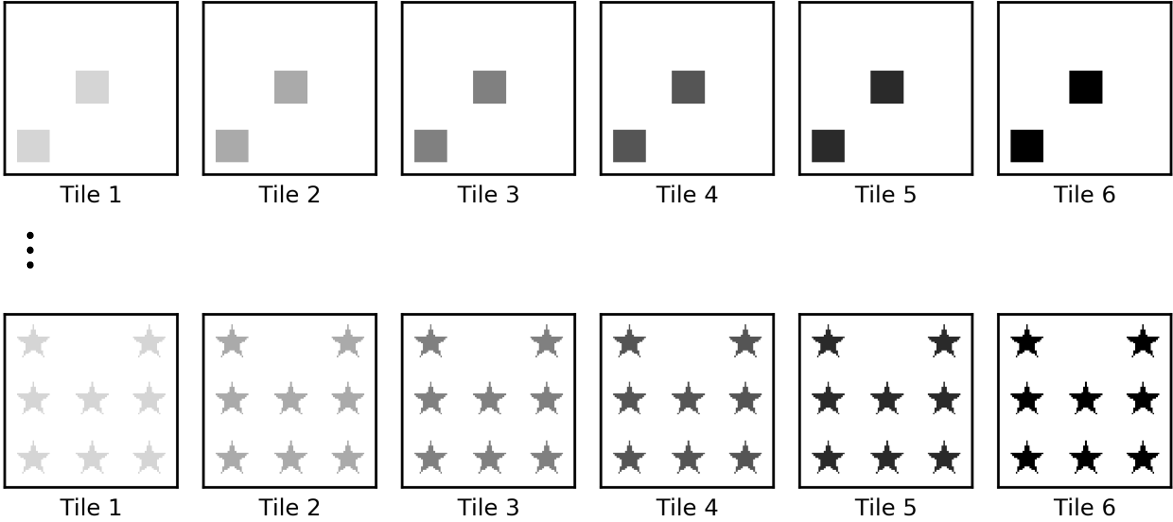}
    \caption{Training set. Rule: color (easy)}
    \label{fig:extensive_a}
  \end{subfigure}
  \hfill
  \begin{subfigure}[b]{0.45\textwidth}
    \includegraphics[width=\textwidth]{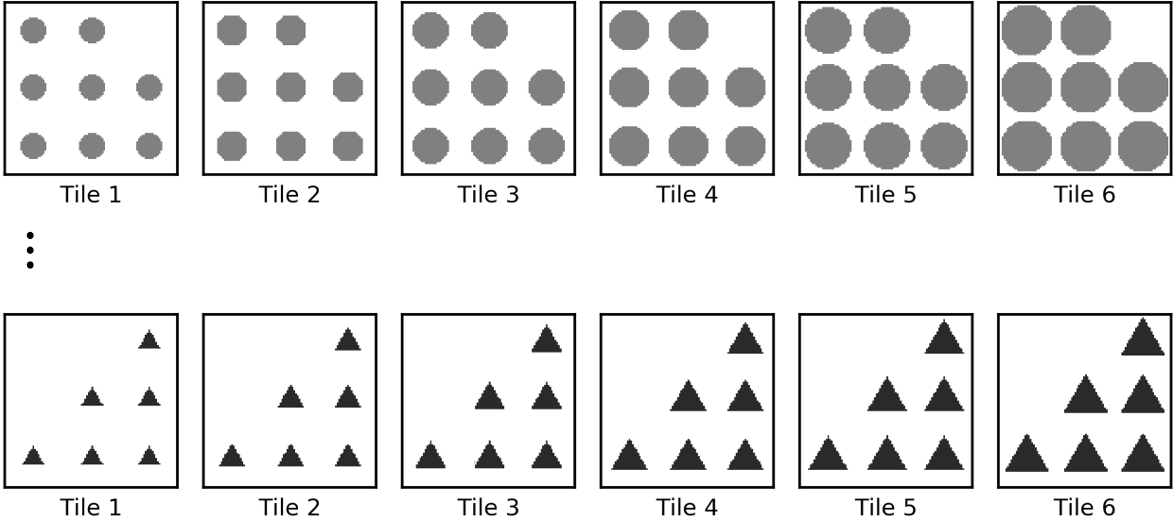}
    \caption{Training set. Rule: size (easy)}
    \label{fig:extensive_b}
  \end{subfigure}
  \\[2ex]
  \begin{subfigure}[b]{0.45\textwidth}
    \includegraphics[width=\textwidth]{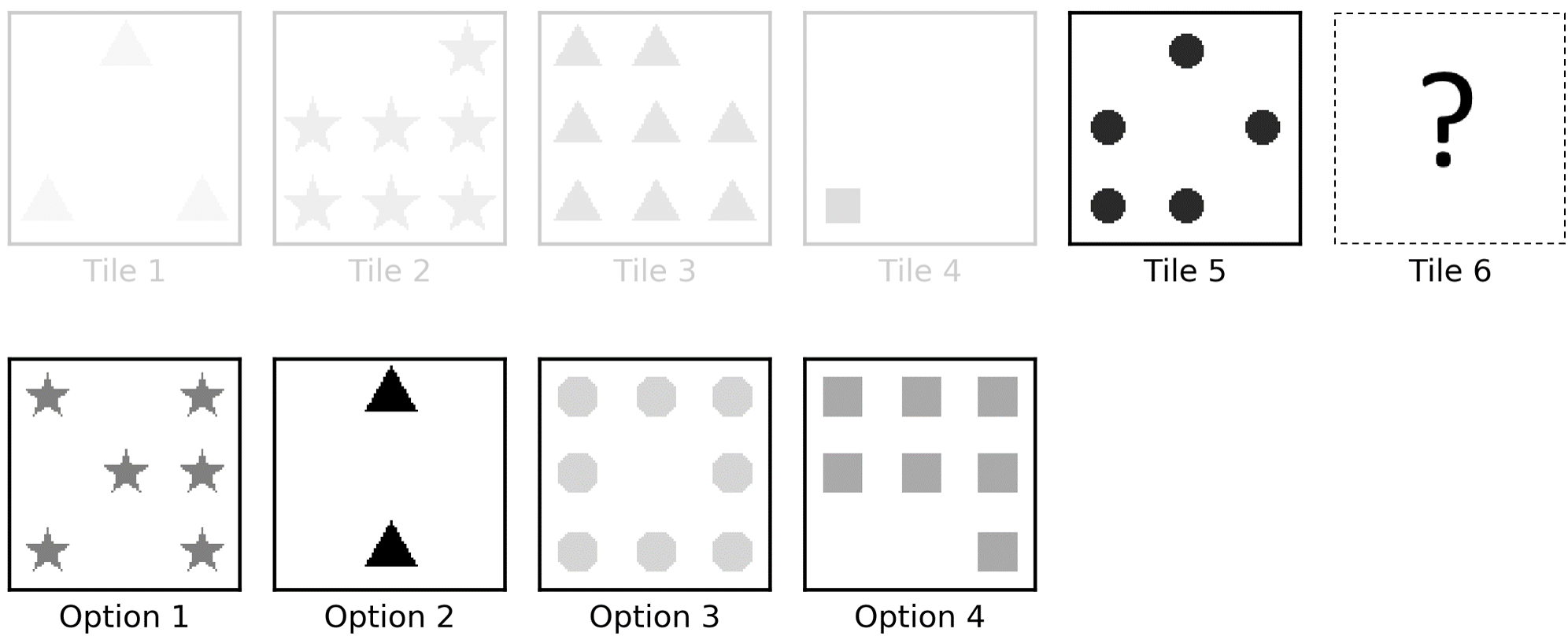}
    \caption{Test set. Rule: color (difficult)}
    \label{fig:extensive_c}
  \end{subfigure}
  \hfill
  \begin{subfigure}[b]{0.45\textwidth}
    \includegraphics[width=\textwidth]{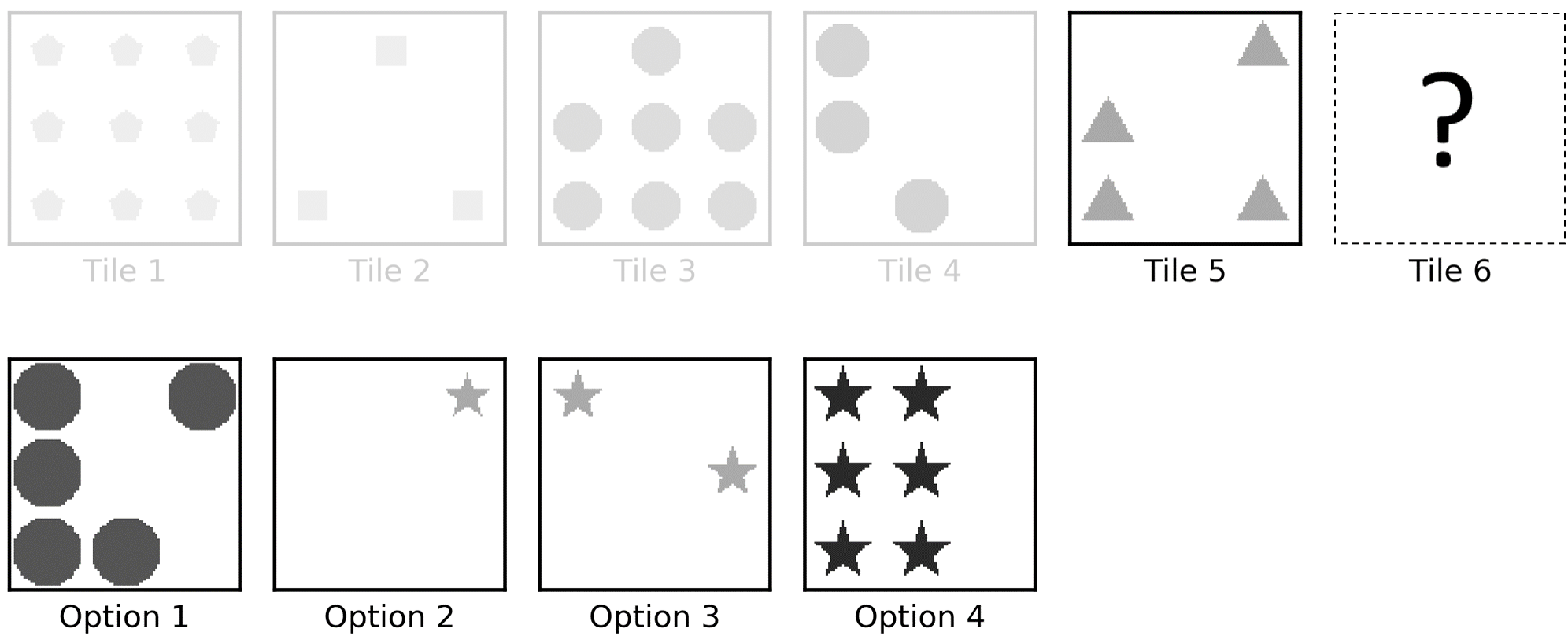}
    \caption{Test set. Rule: size (difficult)}
    \label{fig:extensive_d}
  \end{subfigure}
  \\[2ex]
  \begin{subfigure}[b]{0.45\textwidth}
    \includegraphics[width=\textwidth]{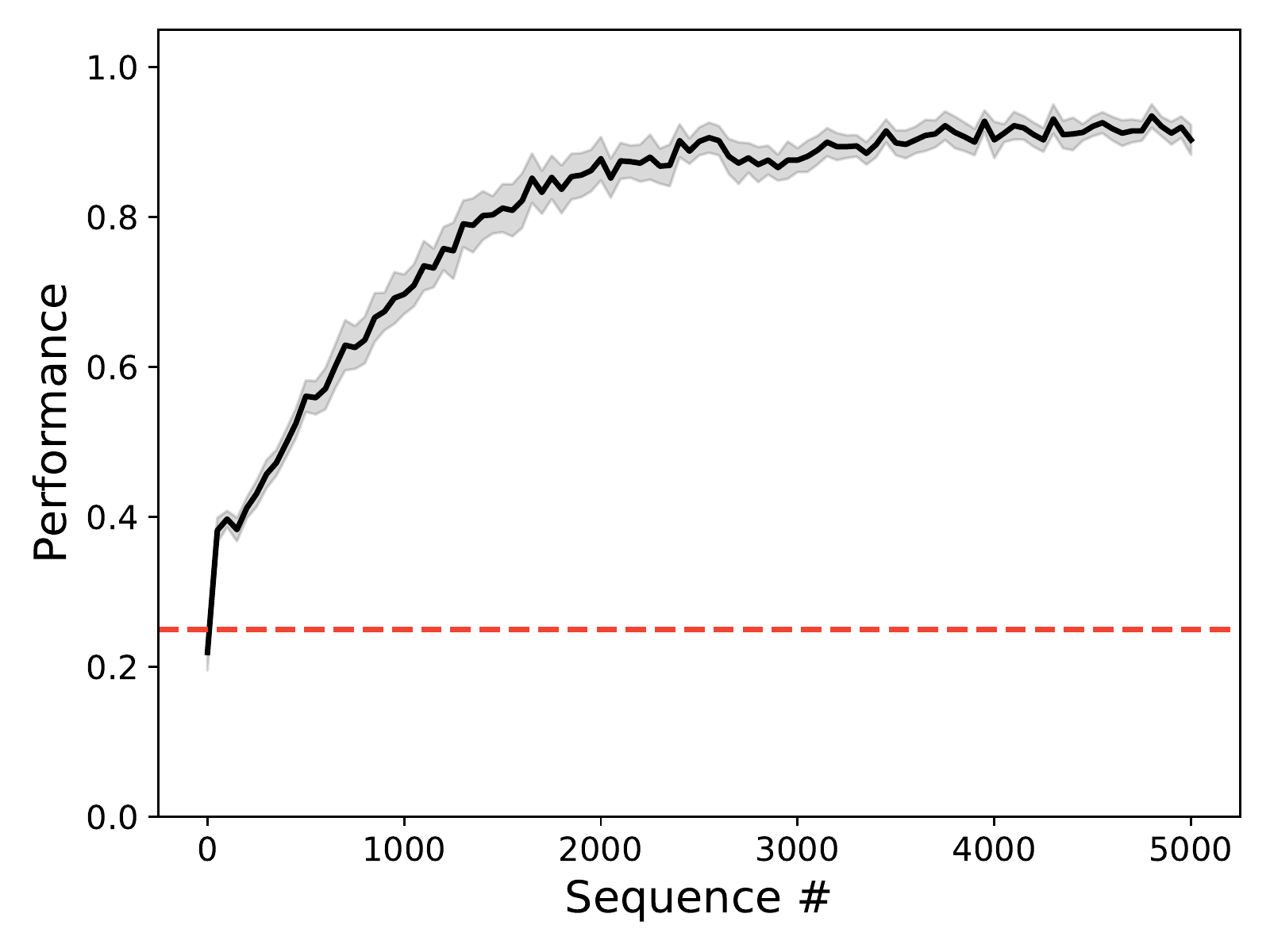}
    \caption{Performance. Rule: color (difficult)}
    \label{fig:extensive_e}
  \end{subfigure}
  \hfill
  \begin{subfigure}[b]{0.45\textwidth}
    \includegraphics[width=\textwidth]{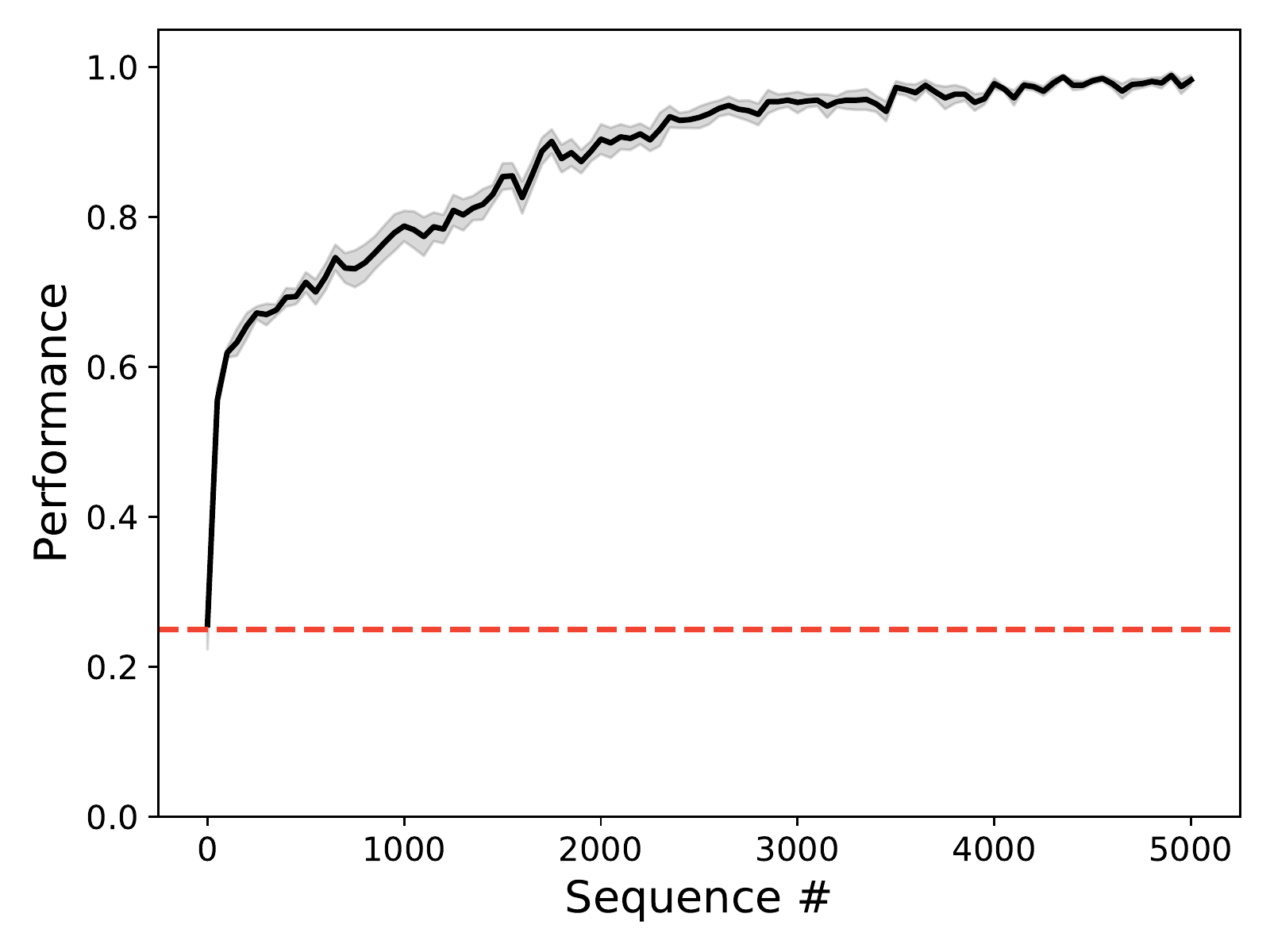}
    \caption{Performance. Rule: size (difficult)}
    \label{fig:extensive_f}
  \end{subfigure}
  \caption{\textbf{Extensive training}. We tested the model's ability to solve our most difficult tests in two test conditions: when the predictable feature is the color of the objects (left side of the figure) versus the size of the objects (right side of the figure). \textbf{\subref{fig:extensive_a},  \subref{fig:extensive_b} Training.} In each condition, 10 networks were extensively trained on easy sequences in which either the size \subref{fig:extensive_a} or the color \subref{fig:extensive_b} were predictable. The networks performed two optimization steps per training sequence, and then moved to the next sequence. \textbf{\subref{fig:extensive_c},  \subref{fig:extensive_d} Test set.} While the networks were training, we measured their performance on 100 difficult tests complying with matching rules (size \subref{fig:extensive_c}, color  \subref{fig:extensive_d}). Note that because the network is trained on sequences other than the test sequence and because prediction in the test sequence relies on the last tile, network's performance is independent of the all test tiles but the last, and hence these tiles are bleached in the figure. \textbf{\subref{fig:extensive_e},  \subref{fig:extensive_f} Performance.} The networks performance in the difficult trials improved with training, reaching success rates that exceeded 90\% in the most difficult tests after thousands of training sequences. Note the fast and substantial improvement after only a few training sequences. The dark lines denote the mean performance and the shades are the standard error of the means (SEMs). The dashed red lines denote the chance levels (25\%).}
  \label{fig:extensive}
\end{figure}

The networks' ability to solve difficult intelligence tests such as the one depicted in Fig. \ref{fig:test_example} depends on it being able to find accurate approximations of $Z^*$ and $T^*$. Specifically, the networks should be sufficiently expressive to approximate $Z^*$ and $T^*$ well enough, and the SGD-based optimization process on the loss functions should converge to such a solution. To test these, we extensively trained the networks on \textit{easy} sequences (Fig. \ref{fig:extensive}a-\ref{fig:extensive}b), in which one feature is monotonically increasing whereas the other features remain constant, and tested them on the \textit{difficult} intelligence tests, in which the predictable feature of the training set followed the same rule but all other features were distractors (i.e., randomly changed) (Fig. \ref{fig:extensive}c-\ref{fig:extensive}d). Training on easy sequences and testing on difficult ones minimized the possibility that a consecutive pair of tiles appearing in the training set would reappear in the test set, thus minimizing the possibility that overfitting underlay our results. Interestingly, we found that training on difficult problems resulted in slower and less robust learning - a further indication that the performance of the networks after learning did not reflect overfitting.

The results of this training procedure are depicted in Fig. \ref{fig:extensive}e-\ref{fig:extensive}f. Within thousands of training sequences, the model achieved success rates that exceeded 90\% (compared with 25\% chance performance), demonstrating that our network is expressive enough to solve the intelligence tests of Fig. \ref{fig:test_example}. The success of the training also indicates that the training procedure, i.e. minimization of the loss functions of equation (\ref{eq:losses}) with three RMSprop optimizers can lead to a good approximation of $Z^*$ and $T^*$. This result is consistent with previous studies that demonstrated that unsupervised latent prediction models are capable of learning good abstract representations when extensively trained (\cite{Oord2018,Anand2019,Yan2020}).

\subsection{Na\"{i}ve networks}
\label{sec:Naive}

\begin{figure}[h]
    \includegraphics[width=\textwidth]{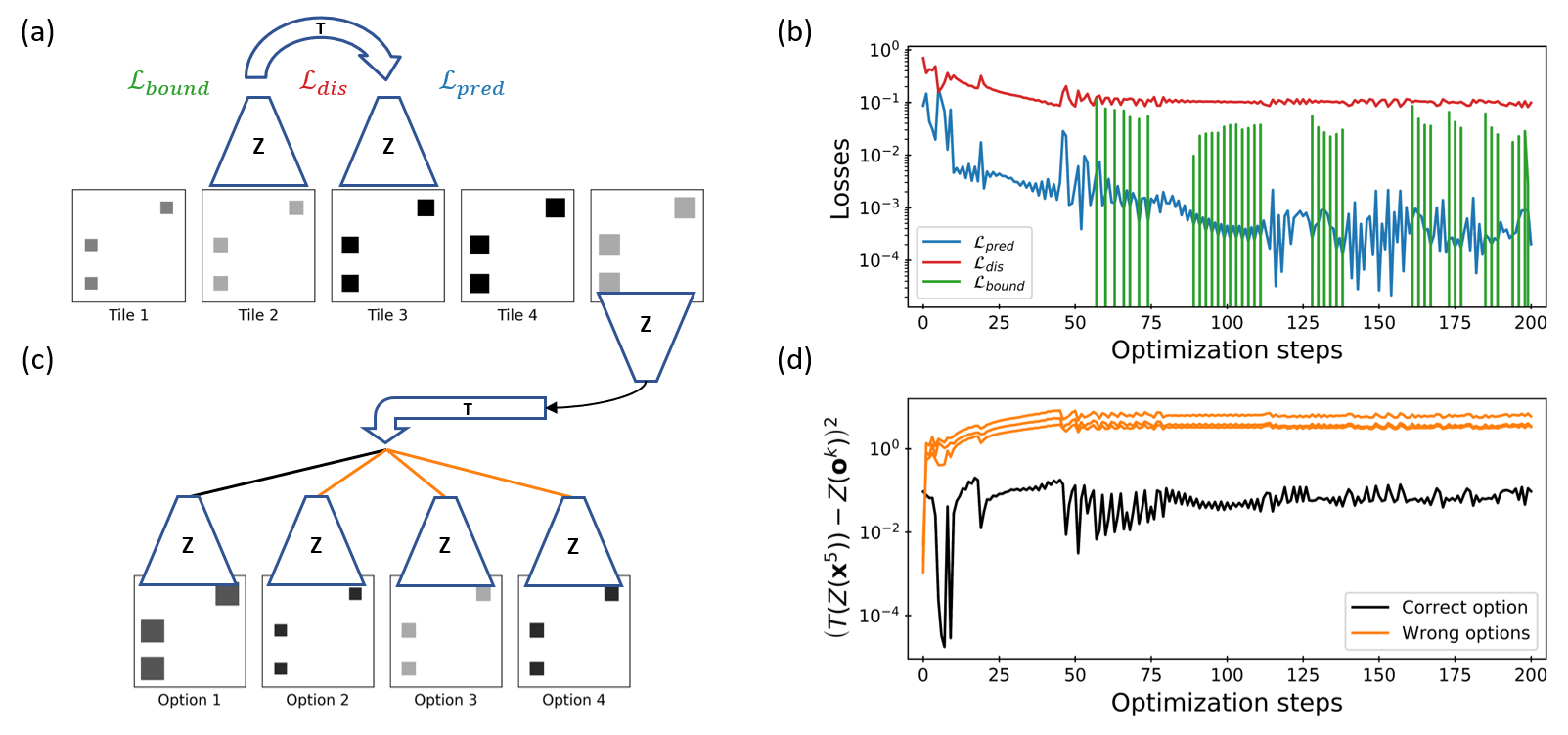}
    \caption{\textbf{The solving of an intelligence test by a na\"{i}ve network}. \textbf{(a)} The network trains on $t=5$ tiles by minimizing the three loss functions. \textbf{(b)} The loss functions as a function of training steps. Note that $\mathcal{L}_{bound}$ occasionally sets the scale of the representations. \textbf{(c)} The prediction errors between the fifth test tile and the $n=4$ options are measured and used for decision-making. \textbf{(d)} The prediction errors signals out the correct option (black) from the incorrect options (orange) after less than 10 optimization steps.}
    \label{fig:naive_example}
\end{figure}

A fundamental difference between the performance of the networks in the previous section and human intelligence is that humans do not seem to require any extensive training in order to solve RPMs (although training does improve performance \cite{Denney1990}). In fact, a hallmark of fluid intelligence is the ability to infer a rule from a very small number of examples. This observation motivated us to study the extent to which our networks can solve intelligence tests in the absence of \textit{any} prior training. To that goal, rather than training using a large number of different sequences of tiles, we used the \textit{test sequence} itself as our training sequence. Fig. \ref{fig:naive_example} depicts an example of a na\"{i}ve network that solve an intelligence test of intermediate difficulty (the test from Fig. \ref{fig:decision_making}). We used exactly the same training procedure as in section \ref{sec:extensive_trainig} with one important difference - we used identical copies of \textit{the same single test sequence} as our training set (Fig. \ref{fig:naive_example}a) The parameters of the encoder $Z$ and predictor $T$ are learned by minimizing the three loss functions over the sequence (Fig. \ref{fig:naive_example}b). Decision was based on the best-predicted option, Eq. \ref{eq:decision} (Fig. \ref{fig:naive_example}c). The prediction errors, $\left( T(Z(\mathbf{x}^t))-Z(\mathbf{o}^k) \right) ^2$ for the correct option $\mathbf{o}^{\text{correct}}$ (black) and incorrect (orange) options as a function of optimization steps are depicted in Fig. \ref{fig:naive_example}d. Within 10 optimization steps, the prediction error associated with the correct tile was already substantially smaller than that of the incorrect tiles, directing choice towards the correct answer.

\begin{figure}[h] 
    \savebox{\imagebox}{\includegraphics[width=0.42\textwidth]{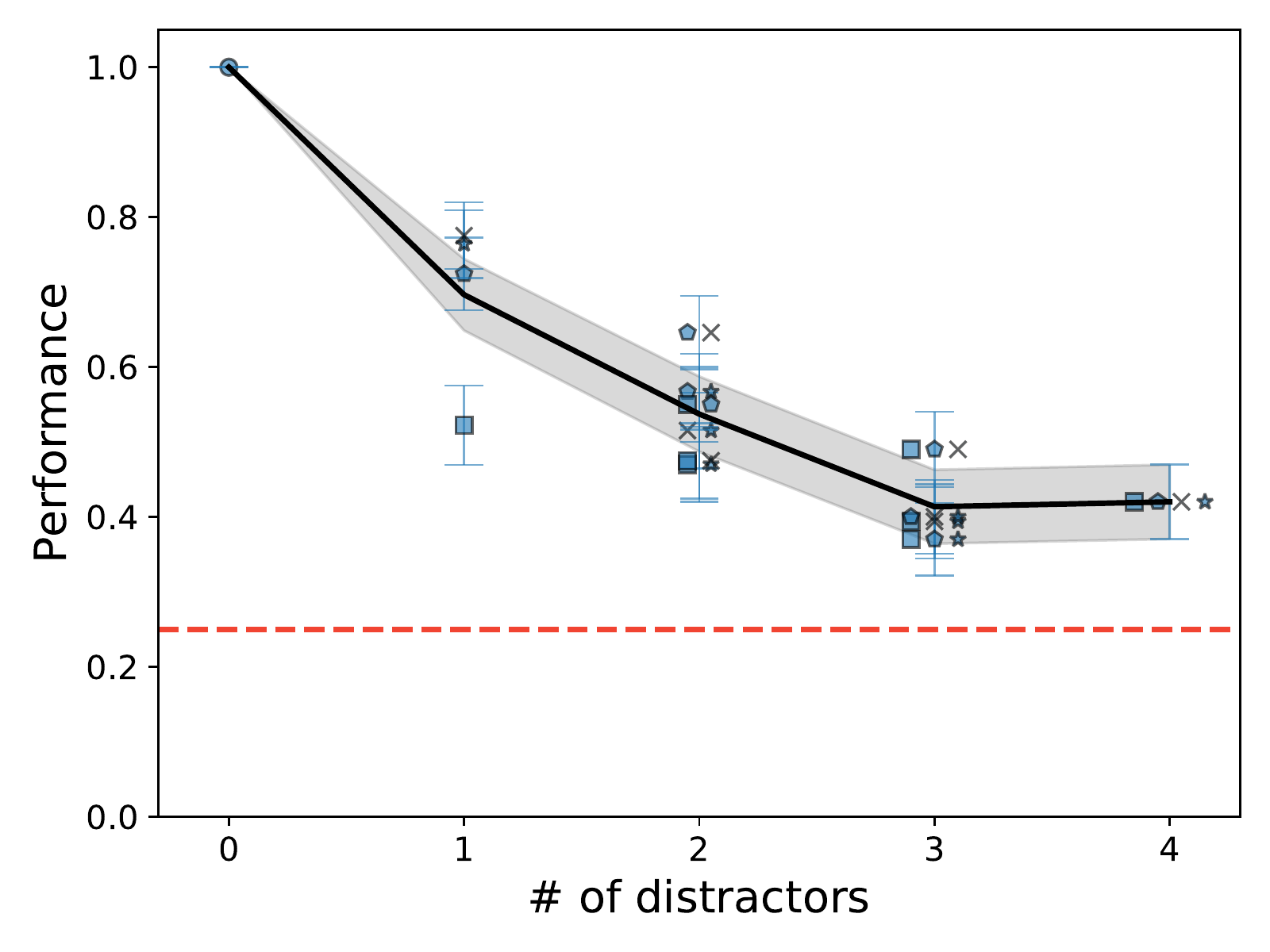}}
    \centering
    \begin{subfigure}[t]{0.42\textwidth}
        \usebox{\imagebox}
        \caption{Rule: color}
        \label{fig:naive_color}
    \end{subfigure}
    \hfill
    \begin{subfigure}[t]{0.10\textwidth}
        \raisebox{\dimexpr\ht\imagebox-\height}{ \includegraphics[width=\textwidth]{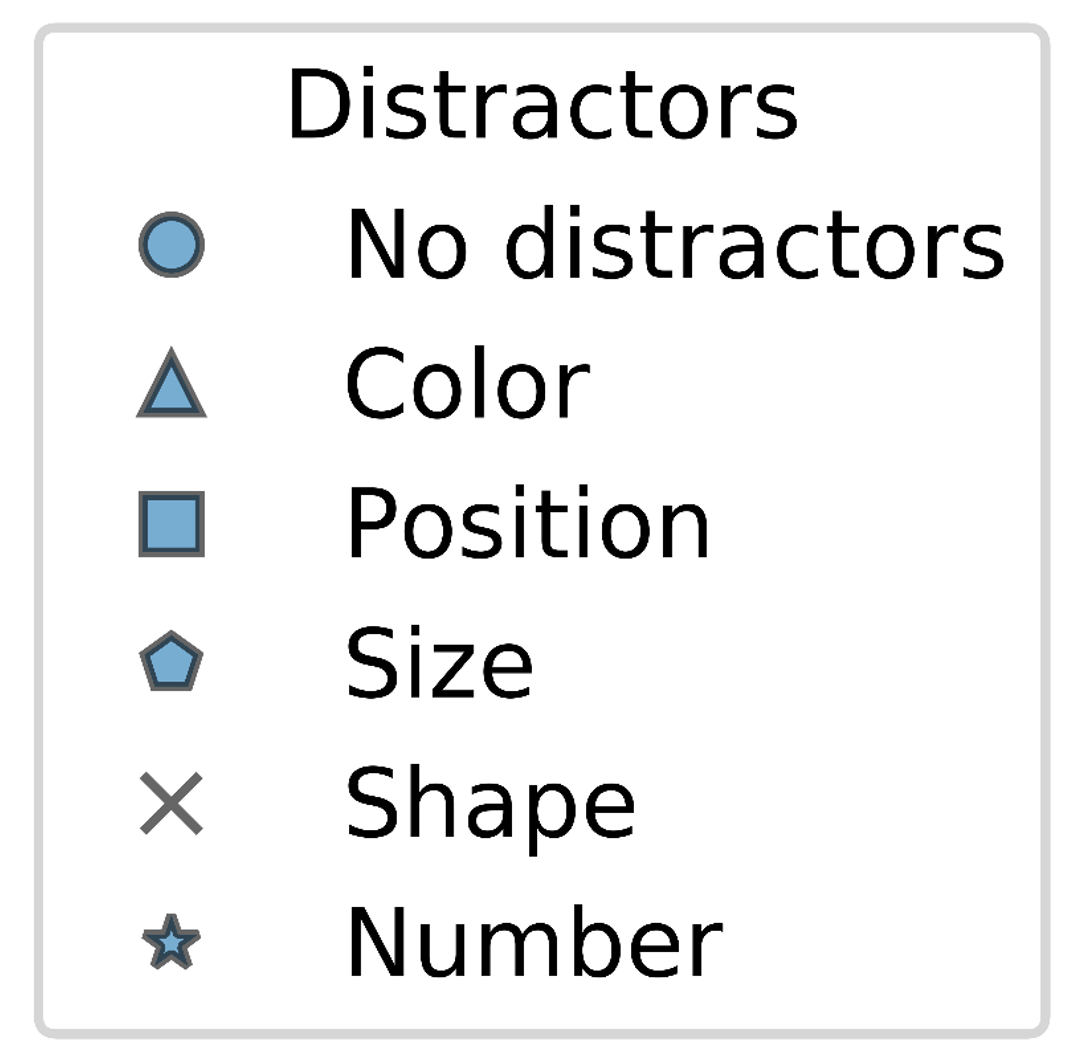}}
    \end{subfigure}
    \hfill
    \begin{subfigure}[t]{0.42\textwidth}
        \includegraphics[width=\textwidth]{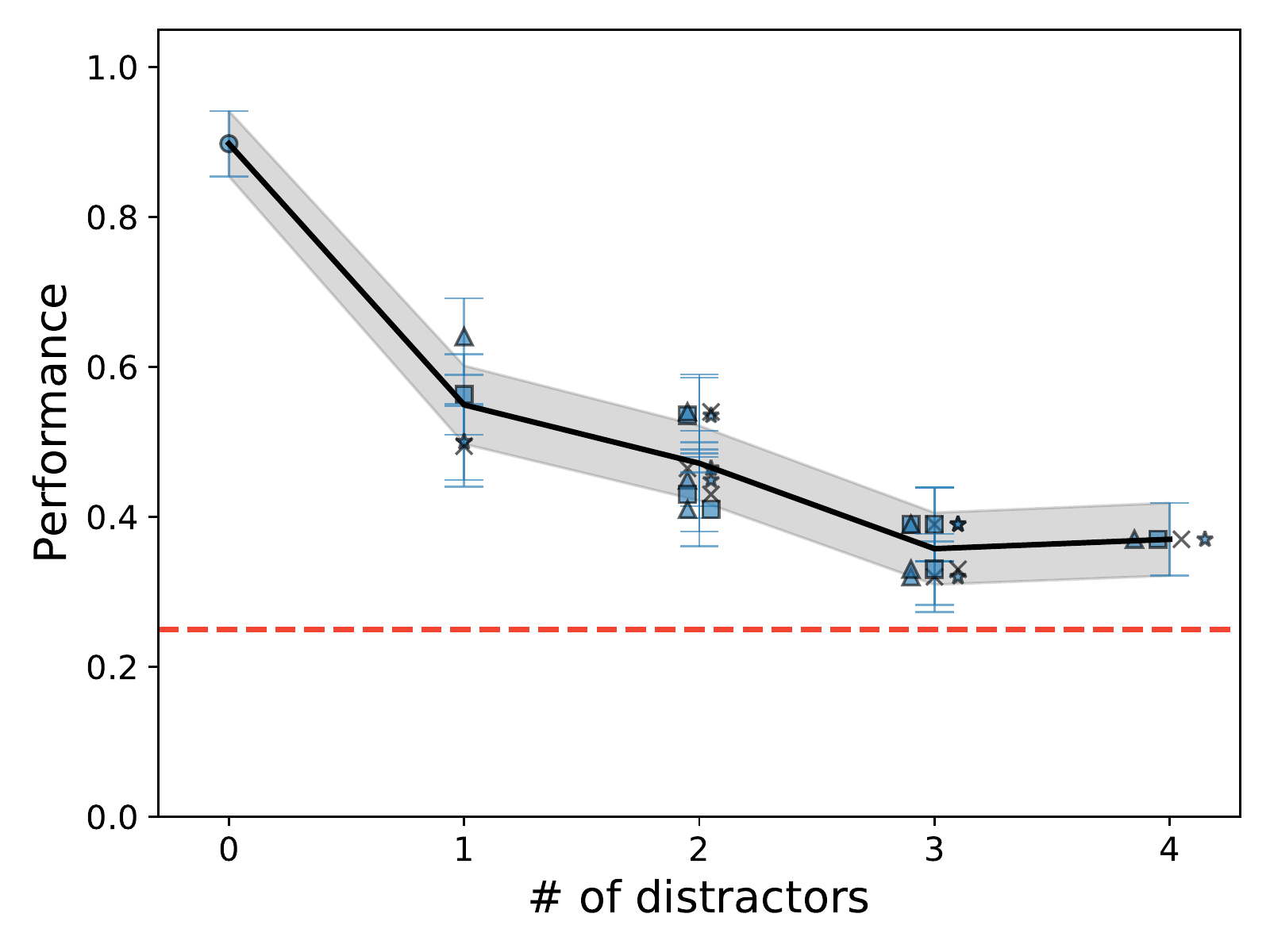}
        \caption{Rule: size}
        \label{fig:naive_size}
    \end{subfigure}
    \caption{\textbf{Na\"{i}ve networks' performance}. The model's ability to solve tests without training was evaluated on multiple test conditions. The test conditions differed in the predictable feature (color in \subref{fig:naive_color} and size in \subref{fig:naive_size}) and by the number and types of distractors: the markers' type denote the type of distractors used (see middle legend). Each test condition was composed of 100 tests and each test was solved by a different randomly initiated network, trained with 200 optimization steps.  The error bars are the standard error of the mean for the corresponding test condition. The black lines denote the mean performance and the shades are the standard deviation over the different conditions.}
    \label{fig:naive_full_results}
\end{figure}

In order to quantitatively quantify the na\"{i}ve networks' performance, we tested the model in multiple test conditions (Fig. \ref{fig:naive_full_results}). Each test condition contained 100 intelligence tests, each solved by a different randomly-initialized network. Remarkably, we found that training on the $t=5$ tiles of the test is sufficient for solving the easy tests, as well as for achieving a level of performance that is substantially higher than chance in the difficult tests. All this was achieved without any prior learning and knowledge, using networks whose weights were randomly-chosen. We posit that the success of the model in solving intelligence tests without any training indicates that the architecture and optimization process can also be used as models for inductive reasoning in the cognitive sciences. 

 \section{Relations to other works}
\label{other_works}

\paragraph{Machine learning and RPMs.} A previous study has shown that with extensive supervised learning, RPMs can be solved by deep networks (WREN \cite{Barrett2018}). Worthwhile noting is that these networks can solve RPMs characterized by rules that cannot be learned by our network, e.g., logical rules that require "working memory". Moreover, learning is faster and performance is improved if latent representations are first learned in an unsupervised way and these representations are then used as input to a supervised-trained network \cite{Steenbrugge2018, VanSteenkiste2019}. The two fundamental differences between these approaches and our na\"{i}ve network are (1) our learning is fully unsupervised and (2) our network does not utilize any prior information beyond the test tiles.  

\paragraph{Unsupervised latent prediction models.} Several studies have proposed to use the \textit{predictive information} between the past and the future for dimensionality reduction. Dynamical Component Analysis (DCA) \cite{Clark2019} finds predictable latent representations $Z(x)$ by maximizing predictive information between the past and future $I(Z(x^\text{past});Z(x^\text{future}))$ using a linear approximation. Another related linear method is Slow Feature Analysis (SFA) \cite{Wiskott2002} that finds slowly varying features of the data. Contrastive predictive coding \cite{Oord2018} is an unsupervised optimization problem that finds such representations using a deep neural network. Such contrastive predictive coding has been successfully used to find useful latent representations of ATARI games \cite{Anand2019} and deformable objects \cite{Yan2020}. Conceptually, our approach is similar to these previous studies. The challenge of finding a solution to the equation $T(Z(x^\text{past}))=Z(x^\text{future})$ such that $T(Z(x^\text{past}))$ is as dissimilar as possible to $Z(x^\text{future})$ can be viewed as an approximation to the challenge of maximizing the predictive information, with the advantage of separating the information into encoding and prediction functions, which is useful for making actual predictions. The ability to make predictions in latent space (world models) has proven useful for planning in RL, which results in improved overall performance \cite{Francois-Lavet2019, Ha2018}. Our latent prediction model is based on these studies, but is used for a very different purpose - a model of fluid intelligence.

\section{Discussion}
\label{discussion}

We identified an analogy between data-efficient latent prediction models and the fluid intelligence's core cognitive ability of inductive reasoning. We used this analogy to build a computational model that can solve fluid intelligence tests without prior training or knowledge.

\paragraph{Data efficiency.} Deep neural networks are expressive enough to overfit large random datasets, and are especially capable of overfitting small number of examples. However, a remarkable feat of deep neural networks is that they can generalize even when the number of examples in the training set is substantially smaller than the number of parameters, a result that is still not fully understood \cite{Zhang2019}. The ability of our networks to approximate a rule by observing only five tiles in the "na\"{i}ve" experiments takes this ability to the extreme. It has been argued that the remarkable capabilities of the human brain to learn from a small number of examples (\cite{Barascud2016}) in comparison to artificial networks result from priors that are learned prior to the experiment, or even in evolutionary time-scales \cite{Chollet2019, Mongillo2014, Mongillo2014a}. Our results indicate that in fact, much can be achieved without priors even when the training set is limited.

\paragraph{The limitations of the model.} There are rules that by construction of the model, cannot be learned. Specifically, rules that require memory, e.g., logical operations and long-term relations between the tiles cannot be learned. Incorporating such rules to the repertoire of the network can be done by defining the input to the encoder to be a set of several consecutive tiles. Alternatively, working memory can be incorporated into the model by replacing the feed-forward networks $Z$ and / or $T$ with recurrent networks \cite{Oord2018}. Another limitation of the model is that the solutions $\hat{Z}$ and $\hat{T}$ are likely to differ from the true solutions $Z^*$ and $T^*$ even when the networks identified the correct solution. Our measure of success is not the learning of the rule, i.e., the similarity between ${\hat{Z}, \hat{T}}$ and ${Z^*, T^*}$. Rather it relies on the ability of the network to choose the correct tile from a finite set of $n=4$ alternatives. The results indicate that the networks found solutions that were correlated with the ground truth, a correlation that enabled them to solve the task.

\paragraph{Fluid vs. crystal intelligence.} We studied the networks' performance in two regimes. Using cognitive terminology, the extensive training experiment was a test of crystal intelligence in which performance improved with the accumulation of knowledge. By contrast, in the na\"{i}ve experiment setting, performance relied on the fluid intelligence of the model - the predefined model architecture, its loss-functions and the optimization process. It could be interesting to combine the two types of intelligence by considering learning in multiple time-scales, the shorter ones corresponding to improving crystal intelligence whereas the longer ones to improving the hyperparameters of the network - hence its fluid intelligence. Improving humans' fluid intelligence via training is a hard challenge in psychology with no existing method showing definite success \cite{TeNijenhuis2007,Au2015,Hayes2015}. Our model puts us in position to try and study the computational requirements for improving fluid intelligence.

\paragraph{Conclusion.} We showed that deep neural networks can solve intelligence tests and exhibit fluid intelligence. Our model demonstrates the potential fluid intelligence of artificial networks and help us identify the computational challenges of fluid intelligence in humans and animals.

\begin{ack}
This work was supported by the Israel Science Foundation (Grant No. 757/16) and the Gatsby Charitable Foundation.
\end{ack}

\bibliography{Main}

\end{document}


\maketitle

\section{Intelligence tests details}

Our method for creating tests is based on \cite{Wang2015,Barrett2018} and goes as follows. A test consists of $t$ sequential images and $n$ optional images for completing the sequence. The images are matrices of $100\times100$ gray-scale pixels. Each image can have up to nine shapes drawn in it, arranged in a $3\times3$ grid. The features that characterize each image are:
\begin{enumerate}
    \item Color - the color intensity of the shapes. The possible values depend on the total length of the sequence, including the correct option, $\{\frac{i}{t+1}\}_{i=1}^{t+1}$.
    \item Positions - a vector that specifies the order in which the shapes are placed in the $3\times3$ grid of the image.
    \item Size - the size of the shapes is defined to be the diameter of their circumscribed circle. The possible values are linearly spaced in the range of $[15,30]$ pixels.
    \item Shape - an image is characterized by one shape type out of five possible types: triangle, square, pentagon, star and circle.
    \item Number - a number from 1 to 9 that specifies the number of shapes in the image.
\end{enumerate}

Initially, each feature is assigned a random value out of its possible values with an equal probability and the first image of the sequence is created. Then, the features change based on the specification of their properties (whether they are constant, random, or follow a rule) creating the remaining images of the test sequence. The correct option of the $n$ alternatives follows the same specification of features, as if it is the $t+1$ image. By contrast, the incorrect options are inconsistent with the rule - while their constant and random features keep their specification, their predictable feature's value is picked at random out of its remaining possible values with an equal probability.

\section{Network model}

The network was adapted from a RL agent that utilized a world model for planning \cite{Francois-Lavet2019}. We removed the RL components and made minor modifications to the remaining unsupervised learning network. The basic elements of our network are the encoder and predictor networks.

\paragraph{Encoder} Convolutional neural network $Z(\mathbf{x})$. Takes an input of $100\times100$ pixels image and encodes it as a scalar output. Network structure:
\begin{enumerate}
    \item Convolutional layer with 16 channels (RelU)
    \item Convolutional layer with 32 channels (RelU)
    \item MaxPool layer
    \item Convolutional layer with 32 channels (RelU)
    \item MaxPool layer
    \item 4-layer fully connected network (tanh); Number of neurons = [200,100,50,10].
    \item Linear fully connected layer
\end{enumerate}

\textbf{Predictor} Fully connected network $T(Z(\mathbf{x}))$. The predictor calculates the transition in the representation $T(Z)=Z+\Delta T(Z)$ where $\Delta T(Z)$ is a trainable network. The network structure of $\Delta T(Z)$:
\begin{enumerate}
    \item 4-layer fully connected network (Relu in the first 3 layers and tanh in the last layer); Number of neurons = [10,30,30,10].
    \item Linear fully connected layer
\end{enumerate}

\paragraph{Learning rates} Each of the three loss functions that are presented in Eq. (5) of the article is minimized with its own RPMprop optimizer. Therefore, each loss function can have a unique learning rate. After a hyperparameters scan, we chose the following learning rates:
\begin{enumerate}
    \item $\mathcal{L}_{pred}$ optimizer: $3\mathrm{e}{-4}$.
    \item $\mathcal{L}_{dis}$ optimizer: $0.7\times 3\mathrm{e}{-4}$.
    \item $\mathcal{L}_{bound}$ optimizer: $3\mathrm{e}{-4}$.
\end{enumerate}

\section{Attached code files}

The supplementary materials contain two python codes:
\begin{enumerate}
    \item \textbf{Naive\_experiment.py} An example of a na\"{i}ve network solving an intelligence test. The code creates a test using Sequential\_RPMs.py, builds a network and optimizes it on the test sequence while measuring the prediction error of the four options. At the end, the model prints the correct option and plots a figure equivalent to Fig. 4 of the article.
    \item \textbf{Sequential\_RPMs.py} a class that creates our intelligence tests. A test is created based on several parameters. Specifically, seq\_prop is a dictionary that specifies the features properties - whether they should be constants, distractors, or change according to a rule (see example at the bottom of the code). The other parameters determine the length of the test sequence, the number of wrong options, the size of the images, and whether to plot and save a figure of the test. After initializing a test object, the images are available as variables of the object, \textit{self.data} is an array of the test sequence images and \textit{self.options} is an array of the options. The correct option is the first image of the array.
\end{enumerate}

\textit{Note: The supplementary materials also contain an image of a question mark (\textbf{question\_mark.png}) that is required for plotting the intelligence tests. Put the file in the same folder as \textbf{Sequential\_RPMs.py}.}

\bibliography{refs}